\documentclass[10pt,twocolumn,letterpaper]{article}

\usepackage{cvpr}              

\usepackage{times}
\usepackage{epsfig}
\usepackage{graphicx}
\usepackage{amsmath}
\usepackage{amssymb}

\usepackage{enumitem}
\usepackage[ruled]{algorithm2e}
\usepackage{array}
\usepackage{multirow}
\usepackage{color}
\usepackage{cases}
\usepackage[export]{adjustbox}
\usepackage[dvipsnames]{xcolor}
\usepackage{textpos}
\usepackage{comment}
\usepackage{textcomp}
\usepackage{listings}
\usepackage{bibentry}
\usepackage[utf8x]{inputenc}
\usepackage{siunitx}

\usepackage{rotating}

\usepackage{booktabs}

\usepackage[pagebackref,breaklinks,colorlinks]{hyperref}

\usepackage[capitalize]{cleveref}
\crefname{section}{Sec.}{Secs.}
\Crefname{section}{Section}{Sections}
\Crefname{table}{Table}{Tables}
\crefname{table}{Tab.}{Tabs.}

\def\cvprPaperID{6} 
\def\confName{CVsports}
\def\confYear{2022}

\begin{document}

\title{SoccerNet-Tracking: Multiple Object Tracking Dataset and Benchmark in Soccer Videos}


\author{
Anthony Cioppa$^1$* 
\quad Silvio Giancola$^2$* 
\quad Adrien Deli{\`e}ge$^1$*  
\quad Le Kang$^3$*
\quad Xin Zhou$^3$*
\and Zhiyu Cheng$^3$
\quad Bernard Ghanem$^2$ 
\quad Marc Van Droogenbroeck$^1$
\and$^1$ University of Liège
\quad $^2$ KAUST
\quad $^3$ Baidu Research
}

\maketitle

\thispagestyle{empty}

\newcommand{\mysection}[1]{\vspace{2pt}\noindent\textbf{#1}}
\newcommand{\Table}[1]{Table~\ref{tab:#1}}
\newcommand{\Figure}[1]{Figure~\ref{fig:#1}}
\newcommand{\Equation}[1]{Equation~\eqref{eq:#1}}
\newcommand{\Equations}[2]{Equations \eqref{eq:#1} and \eqref{eq:#2}}
\newcommand{\Section}[1]{Section~\ref{sec:#1}}
\newcommand{\SoccerNet}{SoccerNet~\cite{giancola2018soccernet}\xspace}
\newcommand{\ActivityNet}{ActivityNet~\cite{caba2015activitynet}\xspace}

\newcommand{\SG}[1]{\textcolor{orange}{[SG:#1]}}
\newcommand{\TODO}[1]{\textcolor{red}{[TODO:#1]}}

\newcommand\blfootnote[1]{%
  \begingroup
  \renewcommand\thefootnote{}\footnote{#1}%
  \addtocounter{footnote}{-1}%
  \endgroup
}

\definecolor{myred}[a=.5]{RGB}{215,25,28} 
\definecolor{myorange}[a=.5]{RGB}{253,174,97}
\definecolor{anthoblue}[a=.5]{RGB}{31,119,180}
\definecolor{anthoorange}[a=.5]{RGB}{255,127,14}
\definecolor{anthogreen}[a=.5]{RGB}{0,150,0}
\definecolor{anthored}[a=.5]{RGB}{150,0,0}
\definecolor{anthobrown}[a=.5]{RGB}{153,76,0}
\definecolor{mygreen}[a=.5]{RGB}{166,217,106} 
\definecolor{mygray}[a=.5]{gray}{0.57}

\definecolor{newanthogreen}[a=.5]{RGB}{101,140,49}
\definecolor{newanthored}[a=.5]{RGB}{191,0,0}
\definecolor{newanthoblue}[a=.5]{RGB}{0,127,255}
\definecolor{newanthogray}[a=.5]{RGB}{76,76,76}

\definecolor{newanthoorangespotting}[a=.5]{RGB}{227,140,16}
\definecolor{newanthobluespotting}[a=.5]{RGB}{31,119,180}
\definecolor{newanthogreenspotting}[a=.5]{RGB}{44,160,44}

\definecolor{newanthoredreplay}[a=.5]{RGB}{183,27,27}
\definecolor{newanthopinkreplay}[a=.5]{RGB}{217,118,213}

\definecolor{newjacobblue}[a=.5]{RGB}{76,114,176}
\definecolor{newjacoborange}[a=.5]{RGB}{221,132,82}

\newcommand{\whitebox}{\hfill\textcolor{white}{\rule[1mm]{1.8mm}{2.8mm}}\hfill}
\newcommand{\redbox}{\hfill\textcolor{myred}{\rule[1mm]{1.8mm}{2.8mm}}\hfill}
\newcommand{\orangebox}{\hfill\textcolor{myorange}{\rule[1mm]{1.8mm}{2.8mm}}\hfill}
\newcommand{\greenbox}{\hfill\textcolor{mygreen}{\rule[1mm]{1.8mm}{2.8mm}}\hfill}
\newcommand{\graybox}{\hfill\textcolor{mygray}{\rule[1mm]{1.8mm}{2.8mm}}\hfill}
\newcommand{\BG}[1]{\textbf{{\color{red}[BG: #1]}}}

\begin{abstract}
Tracking objects in soccer videos is extremely important to gather both player and team statistics, whether it is to estimate the total distance run, the ball possession or the team formation. Video processing can help automating the extraction of those information, without the need of any invasive sensor, hence applicable to any team on any stadium. Yet, the availability of datasets to train learnable models and benchmarks to evaluate methods on a common testbed is very limited. In this work, we propose a novel dataset for multiple object tracking composed of 200 sequences of 30s each, representative of challenging soccer scenarios, and a complete 45-minutes half-time for long-term tracking. The dataset is fully annotated with bounding boxes and tracklet IDs, enabling the training of MOT baselines in the soccer domain and a full benchmarking of those methods on our segregated challenge sets. Our analysis shows that multiple player, referee and ball tracking in soccer videos is far from being solved, with several improvement required in case of fast motion or  in scenarios of severe occlusion.
\blfootnote{\textbf{(*)} Equal contributions. 
Data/code available at \href{https://www.soccer-net.org}{www.soccer-net.org}.
}
\end{abstract}

\section{Introduction}
\label{sec:Intro}

\begin{figure}
    \centering
    \includegraphics[width=\linewidth]{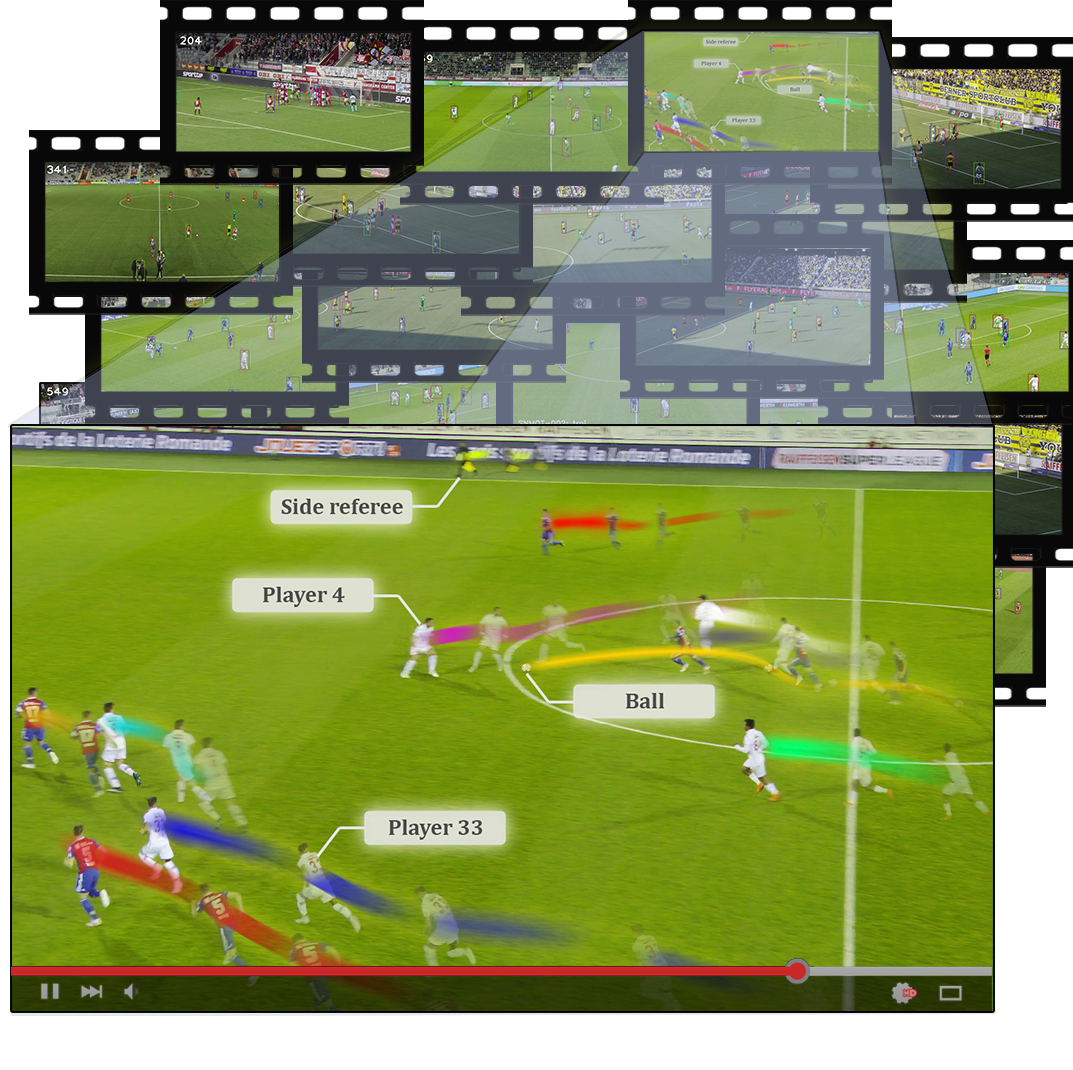}
    \caption{
    \textbf{SoccerNet-Tracking.}
    We propose a novel dataset for Multiple Object Tracking (MOT) in soccer videos including the players, the ball, and the referees. Our dataset is composed of $200$ sequences of $30$s each, representative of interesting moments from $12$ soccer games, densely annotated with player tracklets, teams and jersey numbers. Moreover, we also include a fully annotated $45$min half time video, focusing on long-term tracking.
    }
    \label{fig:graphical_abstract}
\end{figure}


Imagine you are scouting a new striker for your soccer team. How would you evaluate the skills of all potential candidates? 
Prior information on the scouted players are of paramount importance. 
In practice, several metrics typically supports the scouting choices. 
For instance, player endurance along a full game, total distance run, top speed in counter-attacks, number of ball possessions, assists or goals are only few examples of characteristics that your team would consider before hiring your next soccer talent.
Yet, all those important statistics require the information of the players position in time, also known as player tracking.


The increasing demand from sports professionals for more and more advanced analytics calls for automatic video processing.
A low-level understanding of the game and their actors in soccer videos could provide such players and team statistics.
In particular, Multiple Object Tracking (MOT) methods applied to soccer videos could identify the position and trajectory of each player in a video, as well as the ball or any other important actor.
To achieve this objective, visual computing appears handy since it does not require any invasive sensor placed on the players.
Indeed, visual tracking extracts valuable information for each player, such as its positioning in particular scenarios, speed with and without the ball on his feet, running analysis, \etc as those information can indicate a positive outcome such as a goal, for instance. 
Tracking also gathers valuable information for the team as a whole as it helps understanding specific game strategies, and assists in proposing a counter strategy against an adversary team.

However, several challenges arise in soccer player tracking:
\textbf{(i)}~players appear visually very similar between each others in soccer videos, with only a few characteristics to differentiate players from the same team (typically jersey number, or shoes color).
\textbf{(ii)}~Players are often occluding each others in specific game scenarios (\eg corners), increasing the difficulty of recognizing them and with a risk of switching their identities.
\textbf{(iii)}~Tracking the soccer ball is extremely challenging due to its small size (\eg $<100$ pixels), occlusions from players, fast motion, incurring blurring effects, and shape shifting on the video frames.
Yet, several downstream tasks can benefit from soccer player tracking techniques.
For \textit{action spotting}, player trajectories in the field are highly indicative of the nature of the event occurring (\eg crowd around goal when corners happen, players running after a goal occurs, \etc).
For \textit{highlight generation}, an incredible performance from a player can be retrieved by analysing the trajectories of the player and the ball (\eg an impressive sprint or dribbling of several players, a ball crossing the goal line or going out the field, \etc).

In this work, we propose SoccerNet-Tracking, a large-scale dataset and benchmark for Multiple Object Tracking (MOT) in soccer videos, with a focus on players, balls and referees, as illustrated in Figure~\ref{fig:graphical_abstract}. We provide $200$ tracking sequences gathered around $11$ interesting classes of soccer actions (\eg goal, corner, direct free-kick, foul, \etc) and corresponding to challenging tracking scenario. 
Furthermore, we have annotated a complete $45$-minute half-time of densely annotated soccer video to evaluate long-term tracking.
We dedicate part of the sequences for training and benchmarking purposes; annotations for the remaining sequences are kept segregated for future open challenges to prevent any over-fitting.
Finally, we benchmark state-of-the-art MOT baselines on our novel dataset, and run an extensive analysis of challenging tracking scenarios.








\mysection{Contributions.} We summarize our contributions as follows. 
\textbf{(i)} We propose the largest dataset for multi-object tracking in soccer videos. It is composed of $200$ sequences of $30$s each, fully annotated with bounding boxes and ID tracklets at $25$fps, and a complete $45$-minutes half-time for long-term tracking.
\textbf{(ii)} We propose an extensive benchmark of the most recent multi-object trackers on our new dataset, highlighting the difficulties of soccer players tracking in different scenarios, and providing a first state of the art for the tracking task.
\section{Related Work}
\label{sec:RelatedWork}

Our work relates to datasets and methods for multiple object tracking and its application to sports analysis.

\subsection{Multiple Object Tracking (MOT)}

Tracking objects in videos consists in localizing and following objects of interest across video sequences~\cite{Luo2021Multiple}. 

\mysection{Methods.} The task of Multiple Object Tracking (MOT) is often approached with a tracking-by-detection paradigm~\cite{Bochinski2018Extending,Sun2019Deep,Bergmann2019Tracking,Zhang2021Fairmot,Zhang2021Bytetrack}. It consists in localizing objects of interest at each frame, and associating them into tracklets by finding consecutive correspondences in time. Object detection boasts a large active research community; as a result, the research interests in MOT lie in identifying object tracklets out of those detections, handling object disappearances with re-identification techniques and detector failures with temporal interpolation. 
Bewley~\etal~\cite{Bewley2016SimpleOA} proposed \textbf{SORT} that leverages Kalman filtering with an Hungarian algorithm to associate overlapping bounding boxes.
The extension \textbf{DeepSORT}~\cite{DeepSORTpaddleImpl} incorporates deep appearance features into the association metric.
Bergmann~\etal~\cite{Bergmann2019Tracking} proposed \textbf{Tracktor}, a method that exploits the regression head of object detection models to perform temporal realignment.
Zhang~\etal~\cite{Zhang2021Fairmot} proposed \textbf{FairMOT}, that fine-tunes the detection model aside with the re-identification module that associates new detected objects with the previous list of tracklets.
Zhang~\etal~\cite{Zhang2021Bytetrack} proposed \textbf{ByteTrack}, that considers every detection boxes despite their confidence score, relying on Kalman filtering for the association task.
In this work, we apply the latest research in MOT to soccer videos.


\mysection{Datasets.}
The progress in MOT would not have been possible without the numerous datasets released publicly to the community.
The first attempts in building MOT datasets surged from PETS2009~\cite{Ferryman2009Pets2009} and TUD~\cite{Andriluka2010Monocular}, yet they are relatively small in scale.
MOT~\cite{Leal2015MOTchallenge,Milan2016MOT16,Dendorfer2020MOT20}, KITTI-T~\cite{Geiger2012AreWe} and DETRAC~\cite{Lyu2018UA-DETRAC,Wen2020UA-DETRAC} made tremendous efforts in providing curated video sequences annotated with bounding boxes associated in tracklets for the training and evaluation of MOT methods. Those datasets contain different levels of difficulty, from challenging illumination to several occlusions.
Yet, all those datasets focus on tracking pedestrians and/or vehicles for autonomous navigation, traffic monitoring and security purposes. 
Only a few recent works targeted different domains ranging from human faces~\cite{Sundararaman2021Tracking}, animals~\cite{Romero2019idtracker,Pedersen20203D}, biological cells~\cite{Anjum2020CTMC} up to more generic object~\cite{Dave2020Tao}.
In this work, we provide a large scale dataset for MOT in a novel domain, in particular soccer videos.


\subsection{Sports Video Understanding}

Sports videos are investigated for different semantic understandings, ranging from high-level temporal action localization down to low-level player detection and tracking.
Recent years have witnessed a surge in video understanding, with datasets and tasks ranging from 
video classification~\cite{Saraogi2016Event,Khan2018Learning,Gao2020Automatic}, 
action detection~\cite{Tavassolipour2014Event,Sigari2016AFramework,Jiang2016Automatic,Liu2017Soccer,Giancola2018SoccerNet,Fakhar2019Event},
camera shot segmentation~\cite{Deliege2021SoccerNetv2,Jiang2020SoccerDB}, and 
highlight summarization~\cite{Cioppa2020AContextAware}.
%
The SoccerNet series~\cite{Giancola2018SoccerNet,Deliege2021SoccerNetv2,Cioppa2022SoccerNetv3} provides the largest datasets for soccer broadcast understanding, with comprehensive benchmarks for 
Action Spotting, Replay Grounding, Player Re-Identification and Pitch Localization. 
In this work, we complement the SoccerNet effort with the task of multiple object tracking. We propose a novel dataset of $200$ tracking sequences and a complete $45$-minutes half-time for long-term tracking, fully annotated with tracklets of players, referees and balls, meeting the SoccerNet standards in terms of data distribution and benchmark.



\subsection{Player Localization and Tracking in Sports.}
Sports videos are also investigated at a player level, \eg
retrieving player's jersey~\cite{Gerke2017Soccer,Li2018Jersey,Liu2019Pose} or team~\cite{Istasse2019Associative,Koshkina2021Contrastive},
localizing their position in the field~\cite{Rao2015ANovel,Yang20183DMultiview,Sah2019Evaluation}, 
estimating their motion~\cite{Manafifard2017ASurvey} or forecasting their intention~\cite{Theagarajan2018Soccer,Sangesa2020UsingPB}.

\mysection{Player Localization.}
In this literature, Rao~\etal~\cite{Rao2015ANovel} presented a pre-deep learning algorithm that detects players using color gradients and ground lines detected with Hough transform.
Follow up works adapted the latest advances in generic object detection~\cite{Ren2017Faster} for soccer broadcast understanding.
Nekoui~\etal~\cite{Nekoui2020Falcons} investigated sports athletes under challenging positions and Liu~\etal~\cite{Liu2021Detecting} improved object localization by learning their relationships.
Istasse~\etal~\cite{Istasse2019Associative} and Koshkina~\etal~\cite{Koshkina2021Contrastive} learned to discriminate player teams in unsupervised fashions.
Cioppa~\etal~\cite{Cioppa2019ARTHuS} distill, in an online fashion, a segmentation network trained on generic data for single camera soccer videos, and a detection network for real-time player detection in amateur sports~\cite{Cioppa2020Multimodal}.
Sanford~\etal~\cite{Sanford2020Group} and Cioppa~\etal~\cite{Cioppa2021Camera} both leverage player localization for action recognition and spotting.
Inhere, we tackle localization in time, \ie tracking.








\mysection{Player Tracking.}
Tracking players in time is extremely useful to gather per-player statistics. 
The literature in MOT is extensive. Manafifard~\etal~\cite{Manafifard2017ASurvey} provide a comprehensive survey on player tracking in soccer videos.
Iwase~\etal~\cite{Iwase2004Parallel} proposed an elegant solution to track players using a background subtraction method, a triangulation technique from multiple cameras to project their positions on the pitch, and a Kalman filter to identify their trajectories.
Figueroa~\etal~\cite{Figueroa2004Tracking} and Sullivan~\etal~\cite{Sullivan2006Tracking} proposed similar pipelines by considered player positions as nodes in graphs, and trajectories as edges between nodes.
Nillius~\etal~\cite{Nillius2006Multi} presented a Bayesian framework to link identities, and Xing~\etal~\cite{xing2010multiple} included a progressive observation modeling process.
Lu~\etal~\cite{lu2013learning} proposed a learning approach to identify and track players in videos, based on handcrafted visual features and Kalman filters.
More recently, Hurault~\etal~\cite{Hurault2020SelfSupervised} transferred an object detection model trained on generic objects, fine-tuned for soccer player detection and tracking in a self-supervised fashion.
Yet, those methods leverage small-scale and private datasets. With this work, we release the largest ever dataset for MOT in sports, enabling a fair benchmark for future development in this community.

\mysection{Dataset.}
Most works in soccer player MOT benchmark their method on small-scale proprietary datasets~\cite{Bertini2005Player,Tran2011Automatic}.
Pettersen~\etal~\cite{Pettersen2014Soccer} shared videos and GPS tracker logs acquired at the \textbf{Aflheim} Stadium in Norway. Yet, they provide the GPS logs and videos for only $3$ games.
The closest effort to our work originated from Yu~\etal~\cite{Yu2018Comprehensive} and includes the \textbf{SSET} dataset~\cite{Feng2020SSET} and the \textbf{BSPT} baseline~\cite{Song2021Distractor}.
SSET is composed of $282$ hours of video, from which $80$ tracking sequences are extracted. Because SSET relies on broadcast videos, they first identify interesting far-away camera shots. As a results their sequences only lasts about $10s$ ($248$ frames at $25$fps). 
Moreover, SSET focuses on Single Object Tracking rather than Multiple Object Tracking, where a single player of interest is tracked, initialized with its bounding box on the first frame.
Differently, our videos originate from single cameras following the action, and our $200$ sequences are longer ($30$ seconds) and handpicked to represent specific scenarios of interest in a soccer game, representative of interesting game events and tracking scenarios.
Furthermore, our densely annotated $45$-minutes half-time sequence is the first public release of long-term tracking data in the sports community.

\section{SoccerNet-Tracking Dataset}
\label{sec:data}

\begin{table*}[t]
    \centering
    \resizebox{\linewidth}{!}{
    \begin{tabular}{l||r|r|r|r||c|c}
Dataset & Sequences & Frames & Tracklets & Bounding boxes & Domain & Task \\\midrule
MOT16~\cite{Milan2016MOT16}   &  $14$ &   $11{,}235$ & $1{,}276$  & $292{,}733$  & Pedestrians & MOT \\
MOT20~\cite{Dendorfer2020MOT20}   &   $8$ &   $13{,}410$ & $3{,}833$  & $2{,}102{,}385$ & Urban (crowded) & MOT \\
KITTI-T~\cite{Geiger2012AreWe} &  $50$ &   $10{,}870$ & $977$   & $65{,}213$   & Autonomous Driving & MOT \\
Head~\cite{Sundararaman2021Tracking}  &   $5$ &    $5{,}723$ & $2{,}965$  & $1{,}086{,}790$ & Pedestrian (heads) & MOT \\
TAO~\cite{Dave2020Tao}        &   $3$ & $4{,}447{,}038$ & $16{,}104$ & $332{,}401$  & Generic & MOT \\
3DZeF-T~\cite{Pedersen20203D} &   $8$ &   $14{,}398$ & $32$    & $86{,}452$  & Fish & 3D MOT \\
CTMC~\cite{Anjum2020CTMC}     &  $86$ &  $152{,}498$ & $2{,}900$  & $2{,}045{,}834$ & Cells & MOT \\
SSET\cite{Feng2020SSET}           &  $80$ &   $12{,}000$ & $80$    & $12{,}000$   & Soccer & SOT \\ \midrule
SN-Tracking (ours)            & $201$ &  $225{,}375$ & $5{,}009$  & $3{,}645{,}661$ & Soccer & MOT \\
    \end{tabular}
    }
    \caption{\textbf{Comparison of SoccerNet-Tracking with other tracking datasets.} Our dataset contains the largest set of bounding boxes and sequences across all tracking dataset, as well as the second most number of frames and unique tracklets. This shows that SoccerNet-tracking is a great dataset for research in tracking. Also, our dataset is the first multi-object tracking dataset in soccer, scaling by a large factor the previous SSET~\cite{Feng2020SSET} dataset that only focused on single-object tracking.}
    \label{tab:MOT_Comparison}
\end{table*}


\begin{figure}
    \centering
    \includegraphics[width=\linewidth]{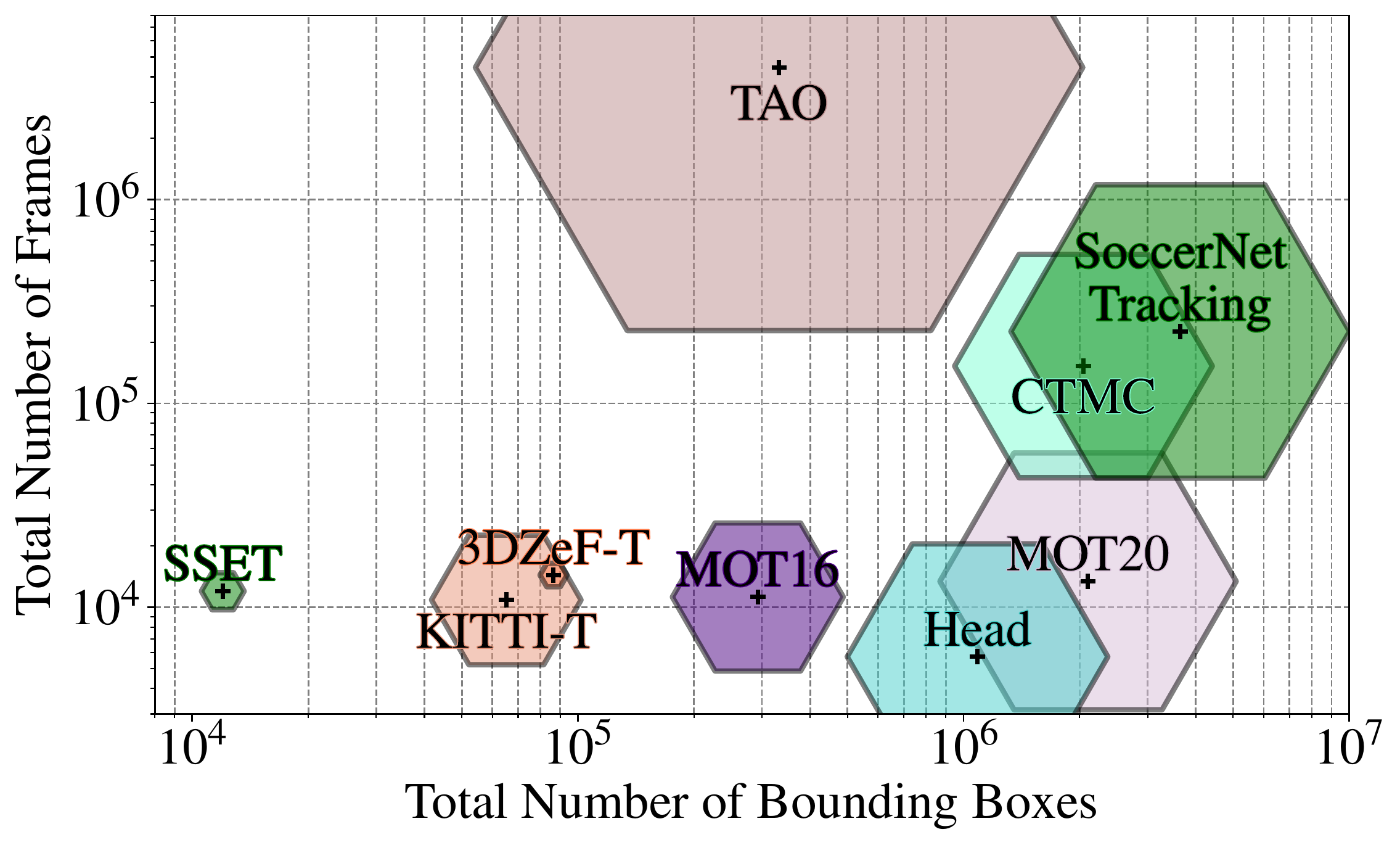}
    \caption{
    \textbf{SoccerNet-Tracking against other tracking datasets.} 
    The area is proportional to the number of unique tracklets in each dataset.
    SoccerNet-Tracking offers a great trade-off between the total number of bounding boxes, frames and unique tracklets.
    Furthermore, only SSET proposes soccer sequences, which is much smaller and only focuses on single object tracking.
    }
    \label{fig:comparison_datasets}
\end{figure}

\begin{figure}
    \centering
    \includegraphics[width=\linewidth]{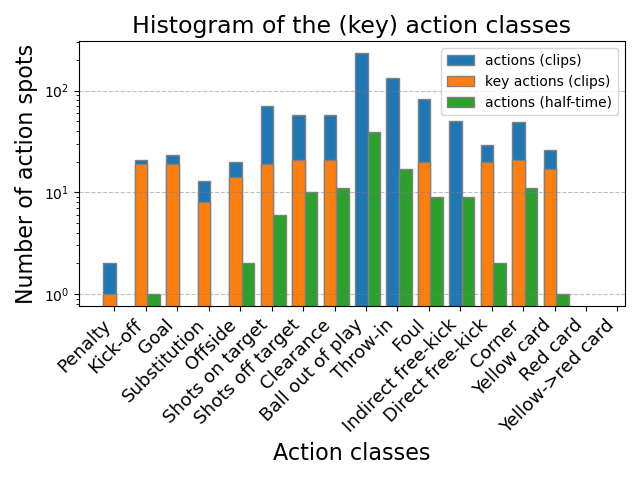}
    \caption{\textbf{Distribution of the action classes.} Number of action classes within the $200$ clips and the whole half-time separately. For the $200$ clips, the key action distribution correspond to the anchoring actions in the clip selection process. Note that within the $12$ games, we have no red card or yellow to red card events.}
    \label{fig:class_distribution}
\end{figure}

\mysection{Data collection.}
Our SoccerNet-Tracking sequences consist of main-camera videos from $12$ complete soccer games recorded during the $2019$ Swiss Super League.
All videos are captured at $1080p$ resolution (Full-HD) and provided at $25$ frames per second.
These single-camera sequences are particularly suited for a tracking task compared with broadcast videos typically found in soccer datasets, where camera changes and replays break the continuity of the tracks.
Since manually annotating the entire $12$ games for tracking would be too costly, we select a subset of these games by extracting $30$-seconds clips at interesting moments in the games, with an additional complete half-time to be annotated.
The first step is therefore to find these interesting moments in the videos.
For that, we start by annotating all events occurring in the game following the same process and action classes than the action spotting task of SoccerNet-v2\cite{Deliege2021SoccerNetv2}.
This resulted in $3{,}132$ extracted action spots corresponding to $17$ action classes among $1{,}210$ minutes of video. 
We noticed that some classes such as for corners or cards are particularly challenging for a tracking task as they often involve clustered players with a lot of occlusions.
Based on these annotations, we select the entire half-time to annotate as the one containing the most challenging action spots, while keeping good action diversity.
Then, among the remaining $11$ games, we select $200$ $30$-seconds clips around key action spots, corresponding to challenging events for tracking, uniformly sampled across $11$ action classes.
Of course, more than one action may happen within these 30-seconds clips, bringing diversity in player configurations and movements.
Finally, we ensure that two clips never overlap to avoid redundancy.
All-in-all, the selected data amounts to $150{,}000$ frames for the clips and $75{,}375$ frames for the whole half-time, totalling $225{,}375$ frames annotated with tracking information, as shown in Table~\ref{tab:MOT_Comparison} and illustrated in Figure~\ref{fig:comparison_datasets}.
The action class distribution among the video clips and the half-time is given in Figure~\ref{fig:class_distribution}. 

\begin{figure*}[t]
    \centering
    \includegraphics[width=\textwidth]{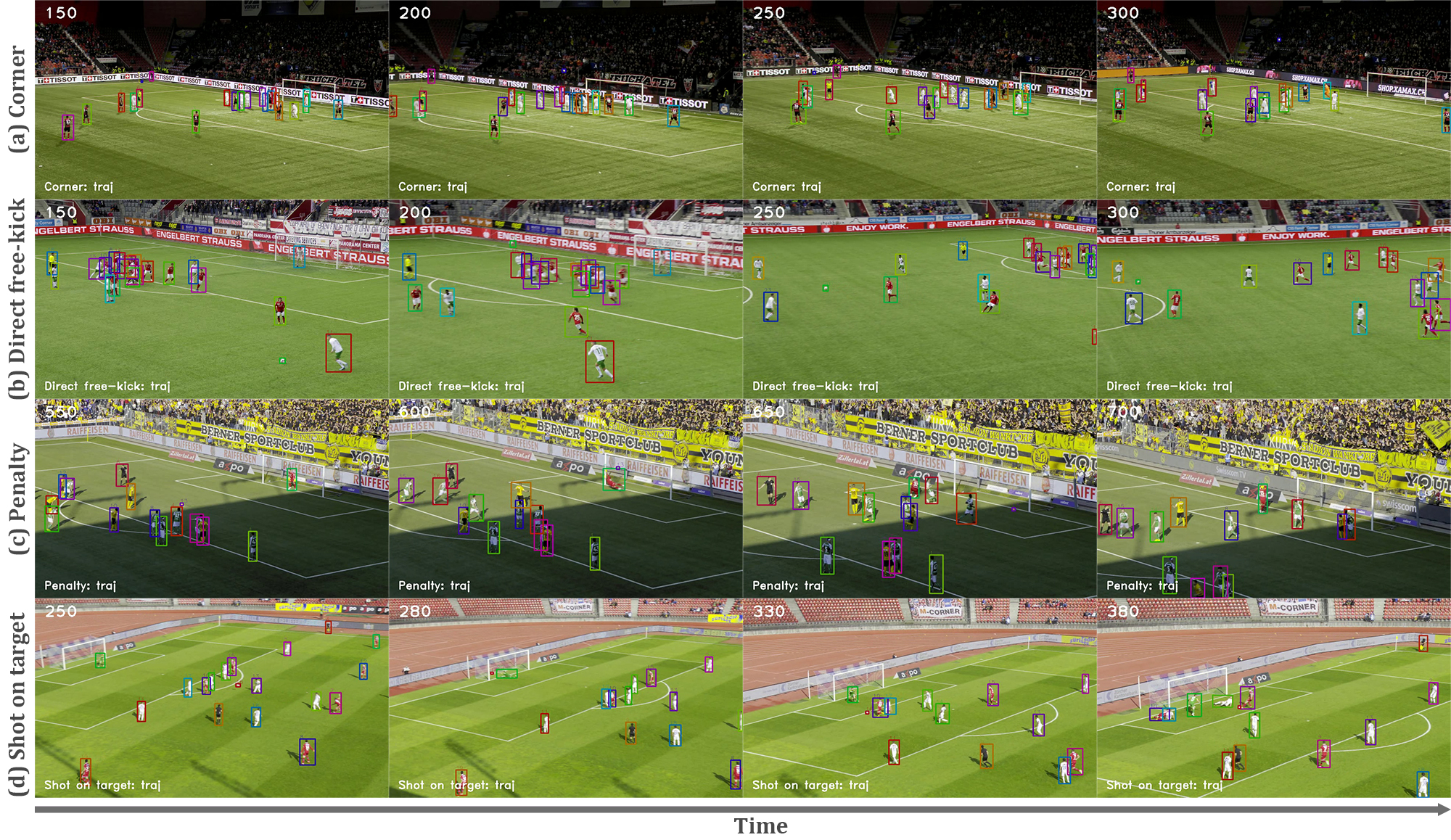}
    \caption{\textbf{Example of tracking annotations in our dataset for challenging events.} From top to bottom: (a)~Corner actions often display a lot of occluded and clustered players. (b)~Direct free-kicks also show clustered players going in the same direction with many crossings between players. (c)~Penalties display almost all objects moving in the same direction often followed by cheering. (d)~Shot on target actions involve high speed movement of the ball and players towards the goal.}
    \label{fig:examples}
\end{figure*}

\mysection{Tracking annotations.}
The tracking information is annotated on SuperAnnotate~\cite{superannotate}, a professional platform specialized in data annotation.
We first define $5$ classes of objects of interest to track, corresponding to the main actors of a soccer game: player, goalkeeper, referee, ball, and other (\eg medical staff or coaches entering the field). 
For the players and goalkeepers, we annotate an extra team tag specifying the side of the team (left or right) as well as the jersey number when visible at least once in the video.
We also further refine the referee annotation between main referee, side top referee and side bottom referee.
Even though these extra metadata are not used for the tracking task described in this paper, they might be useful for further work on team assignment and jersey number recognition.
All bounding boxes are manually annotated as tight as possible around the object of interest at specific key frames. 
In between key frames, the bounding boxes are interpolated both in their position and size using linear interpolation.
On average, only $11.5\%$ of the bounding box are interpolated in the videos, corresponding to very dense manual annotations.
Finally, the bounding boxes are assigned a unique track identifier (ID). 
In total, we have annotated around $3.6M$ bounding boxes corresponding to $5{,}009$ unique objects, with $96\%$ of players and goalkeepers having a jersey number assigned.
Table~\ref{tab:objects_statistics} provides further statistics about the object class distribution.

\begin{table}
    \centering
    \begin{tabular}{l||r|r}
        Object Class & Unique tracklets & Bounding boxes \\\midrule
        Player & $4{,}005$ & $2{,}992{,}173$ \\
        Goalkeeper & $262$ & $130{,}109$ \\
        Referee & $432$ & $301{,}025$ \\
        Ball & $297$ & $215{,}156$ \\
        Other & $13$ & $7{,}198$ \\\midrule
        \textbf{Total} & $5{,}009$ & $3{,}645{,}661$
    \end{tabular}
    \caption{\textbf{Object statistics.} Players are way more represented both in terms of unique tracklets and bounding boxes than other classes.}
    \label{tab:objects_statistics}
\end{table}

\mysection{Novelty.}
Unlike traditional tracking datasets, we purposely choose to keep the same ID for an object that leaves the camera frame and comes back at a later time during the same clip.
This makes our setup much closer to real-world soccer application, where identifying and tracking a player are key components for analyzing his performances.
Most current trackers do not propose such long-term re-identification, making our dataset a perfect sandbox for pushing research towards long-term object re-identification in tracking.
Furthermore, the players of a same team have very similar appearances, making the task particularly challenging in case of crossings between players. Some examples of tracking annotations may be found in Figure~\ref{fig:examples} for several hard cases, including player clusters and crossing between players.
SoccerNet-Tracking is among the largest tracking datasets publicly available, and the largest tracking dataset related to sports. Table~\ref{tab:MOT_Comparison} and Figure~\ref{fig:comparison_datasets} compare SoccerNet-Tracking with the other tracking datasets. As can be seen, our dataset contains the most bounding boxes and sequences, as well as the second most number of frames and unique tracklets. Furthermore, it is the first MOT dataset in soccer, supplanting the previous SSET~\cite{Feng2020SSET} dataset that only focused on single-object tracking.

\mysection{Data format.}
The 12 games are split equally into 4 sets: train, test, and two challenge sets which are kept private at the moment.
In particular, this accounts for $57$ $30$-seconds clips for the train set, $49$ clips for the test set, $58$ clips for our first public challenge, and $37$ clips for our second challenge, including the entire half-time video in the latter.
Then, the folder and data structure are chosen to be as close as possible to the MOT20~\cite{Dendorfer2020MOT20} format. 
We believe that uniformity between datasets is valuable for the tracking community, especially to benchmark new methods. 
In particular, one can use the evaluation and visualization kits of MOT such as \textit{TrackEval}~\cite{Luiten2020Trackeval} and \textit{MOTChallengeEvalKit}~\cite{Dendorfer2020MOTChallengeEvalKit} on our data.
For the sake of completeness, we detail the data format in the following, highlighting the slight non-disruptive discrepancies with the MOT20 dataset.
Each set is separated in its own folder containing all of its sequences, also split in separate sub-folders named after the sequence name. 
The list of sequences in a particular set may be found in the \textit{seqmaps} folder, following the MOT20 convention.
For each sequence, images are extracted from the video and converted to JPEG files named using the frame ID (for instance from \textit{000001.jpg} to \textit{000750.jpg} for the $30$-seconds clips).
The ground truth and detections are stored in their own sub-folder in comma-separate csv files with $10$ columns. 
These values correspond in order to: frame ID, track ID, top left coordinate of the bounding box, top y coordinate, width, height, confidence score for the detection (always 1. for the ground truth) and the remaining values are set to -1 as they are not used in our dataset, but are needed to comply with the MOT20 requirements.
We also provide two configuration files for each sequence. The first one, \textit{seqinfo.ini}, provides information about the video format, such as the length and resolution. The second configuration file, \textit{gameinfo.ini}, provides information about the events and objects in the video sequence such as the anchoring main event and the position of the clip within the whole game. It also provides metadata about the tracks such as the precise object class and jersey number when available (otherwise, the tag is a generic letter used to uniquely identify the object).
Note that unlike MOT20, we provide the ground-truth annotations for the test set publicly, so that researchers can benchmark their results locally without relying on an external evaluation server. However, the ground-truth data for both challenges are kept private and the evaluation can only be done on our submission platform with limited daily and monthly submissions to prevent overfitting.

\section{Benchmarks}
\label{sec:Benchmarks}

\mysection{Task.}
Multi-object tracking (MOT) aims at recovering trajectories of multiple objects in time by estimating object bounding boxes and identities in videos sequences.
In our case, the objects of interest include players, goalkeepers, referees, balls, and other actors such as the medical staff or the coaches entering the field. 
In this work, we consider two tasks: (1) a pure re-identification task that considers ground-truth detections, or (2) a complete tracking task that expects detecting the objects of interest from the raw video.

\mysection{Baselines.}
We evaluate three state-of-the-art MOT methods.
    \textbf{(i)} \emph{DeepSORT}~\cite{DeepSORTpaddleImpl} is an extension of SORT~\cite{Bewley2016SimpleOA} that performs Kalman filtering~\cite{kalman1960new} on detected objects and applies the Hungarian algorithm with an association metric that measures bounding box overlap. DeepSORT extends SORT by incorporating appearance features generated by a deep convolutional neural network into the association metric.
    \textbf{(ii)} \emph{FairMOT}~\cite{Zhang2021Fairmot} performs both object detection and re-identification feature generation in a single shot, similar to JDE~\cite{wang2019towards}, but aims for a good balance between the two tasks. 
    \textbf{(iii)} \emph{ByteTrack}~\cite{Zhang2021Bytetrack} proposes an effective and generic association method that tracks objects by associating every detection box instead of only the high score ones. For the low score detection boxes, ByteTrack uses IOU scores as their similarities to assign tracklets and recover true objects or filter out the background detections. For the detection part, ByteTrack leverages the new YOLOX~\cite{yolox2021} detector. It is also the current state of the art on the MOT20 dataset.
%

\mysection{Metrics.}
To evaluate the different aspects of tracking, we consider two main metrics: MOTA from the CLEAR metrics~\cite{Bernardin2008Evaluating}, and the more recent metric HOTA~\cite{Luiten2021Hota}.
The MOTA~\cite{Bernardin2008Evaluating} metric has been widely used as the main metric for many MOT benchmarks. However, it focuses more on the detection performance and weighs significantly less on the association performance. 
To circumvent this limitation, Liuten~\etal~\cite{Luiten2021Hota} disentangle the performances for detection (DetA) and association (AssA) and combine both in a single HOTA metric.
More details on the HOTA metric can be found in their paper~\cite{Luiten2021Hota}.
For our benchmark, we consider HOTA as the main metric.
We evaluate the above baseline methods on our SoccerNet-Tracking test set, and show the results in Table~\ref{table:results}. In the ``Setup'' column, ``w/ GT'' indicates that ground-truth detections are provided to the baselines, while ``w/o GT'' indicates the more challenging setting, \ie each algorithm uses its own detector. Therefore, in the ``w/ GT'' setup we are able to focus our study only on the association performance, given the same perfect detection results for every algorithm.


\mysection{Implementation details.}
For both DeepSORT and FairMOT, we use the implementations from PaddleDetection~\cite{ppdet2019}, with input dimension $1088\times608$. Specifically, the DeepSORT model uses ``JDE YOLOv3'' for object detection and ``PCB pyramid'' for extracting re-ID features (see \cite{DeepSORTpaddleImpl} for more details). The JDE YOLOv3 detector is pretrained for 30 epochs on the ``MIX'' dataset, which is a collection of six datasets including Caltech Pedestrian~\cite{dollarCVPR09peds}, CityPersons~\cite{Shanshan2017CVPR}, CUHK-SYSU~\cite{Xiao2016EndtoEndDL}, PRW~\cite{zheng2016person}, ETHZ~\cite{eth_biwi_00534}, and MOT17~\cite{Milan2016MOT16}. The FairMOT model uses the DLA-34 backbone, also pretrained on the same ``MIX'' dataset for $30$ epochs, as described in \cite{FairMOTpaddleImpl}. 
We also fine-tune the pre-trained FairMOT model on our train set for $10$ epochs, and denote the resultant model FairMOT-ft.
For ByteTrack, we use the open source code~\cite{ByteTrackImpl} provided by the authors. We use the pre-trained model ``bytetrack\_x\_mot20'' trained on CrowdHuman \cite{shao2018crowdhuman} and MOT20 \cite{Dendorfer2020MOT20} datasets. The option for mixed precision evaluation (flag fp16) is turned on and ``match\_thresh'' is set to 0.8. To use the pretrained model without ground-truth detections, the images are resized to $1600\times896$. With the ground-truth detections, we keep the original image size of $1920\times1080$. All other parameters are set to their default values.





\begin{table}[t]
    \centering
    \resizebox{\linewidth}{!}{
    \begin{tabular}{l|c||c|c|c||c} 
        Algorithm     & Setup &   HOTA    &   DetA    &   AssA    &   MOTA      \\         \midrule
        DeepSORT & w/ GT       &   69.552  &   82.628  &   58.668  &  \bf  94.844   \\     
         FairMOT & w/ GT        &   -  &    -  &    -  &    -     \\  
        ByteTrack & w/ GT      & \bf 71.500      & \bf 84.342  &	\bf 60.718 & 94.572  \\ \midrule
        DeepSORT & w/o GT      &   36.663  &   40.022  &   33.759  &   33.913  \\       
        FairMOT & w/o GT       &   43.911  &   46.317  &   41.778  &   50.698   \\         
        ByteTrack & w/o GT     & 47.225    &	44.489  &	50.257  &   31.741     \\        \midrule
        FairMOT-ft & w/o GT    &  \bf 57.882  &   \bf 66.565  &  \bf  50.492  &  \bf  83.565    \\          
    \end{tabular}
    }
    \caption{\textbf{Leaderboard}. Evaluation of the state-of-the-art tracking methods on our new SoccerNet-Tracking test set with (w/) and without (w/o) ground-truth detections. DeepSORT, FairMOT and ByteTrack increasingly improve the performance, and fine-tuning FairMOT leads to the first state of the art on our new dataset.}
    \label{table:results}
\end{table}

\mysection{Main Results.}
The results for the three baselines are presented in Table~\ref{table:results} for our two tracking setups, \ie with and without ground-truth detections. 
First, \emph{when ground-truth detections are provided}, ByteTrack slightly outperforms DeepSORT in AssA. 
Yet, they have similar MOTA scores, indicating that both algorithms perform quite similarly on the association task only.
It is important to note that both algorithms do not achieve $100\%$ in DetA, as common intuition would suggest. This is due to the fact that some of the detections get filtered out in the association process, especially in the case of fast motion.
Let us note that we could not test FairMOT in this setup as its pipeline does not allow injecting external detections for the association task.

\begin{table}[t]
    \centering
    \resizebox{\linewidth}{!}{
    \begin{tabular}{l||c|c}
Algorithm & \multicolumn{2}{c}{ByteTrack - HOTA (DetA / AssA)} \\
Setup & w/ GT & w/o GT  \\ 
\midrule
Clearance           &   71.0 (82.2 / 61.7) &   49.6 (48.9 / 54.7) \\
Corner              &\underline{63.9 (83.8 / 49.6)} &   41.6 (45.5 / 39.7) \\
Direct free-kick    &   66.5 (84.5 / 53.2) &   46.2 (55.1 / 39.5) \\
Foul                &\underline{65.7 (82.6 / 53.3)} &   50.4 (54.8 / 46.9) \\
Goal                &   70.9 (82.2 / 61.2) &\underline{40.3 (29.6 / 54.8)} \\
Kick-off            &   69.3 (84.5 / 57.1) &   47.6 (49.8 / 46.3) \\
Offside             &   71.2 (83.3 / 61.1) &   46.3 (42.7 / 53.4) \\
Penalty             &   70.4 (84.1 / 58.9) &\underline{31.0 (20.1 / 47.8)} \\
Shots off target    &   74.0 (84.8 / 65.1) &   49.2 (47.5 / 51.9) \\
Shots on target     &   67.6 (83.3 / 55.4) &   46.6 (51.6 / 43.3) \\
Substitution        &\bf84.2 (90.3 / 79.3) &\bf64.7 (63.1 / 66.3) \\
Yellow card         &\bf74.1 (83.6 / 65.7) &\bf53.5 (54.9 / 52.8) \\
    \end{tabular}
    }
    \caption{\textbf{Per-class analysis.} Comparison of HOTA (DetA/AssA) for different event classes for ByteTrack w/ and w/o ground truth detections. 
    The \underline{hardest events} include goals, penalties, and corners with many clustered players, while the \textbf{easiest events} correspond to substitutions and yellow card with mostly static players.}
    \label{table:hota_categories}
\end{table}

\begin{figure*}[t]
    \centering
    \includegraphics[width=.97\textwidth]{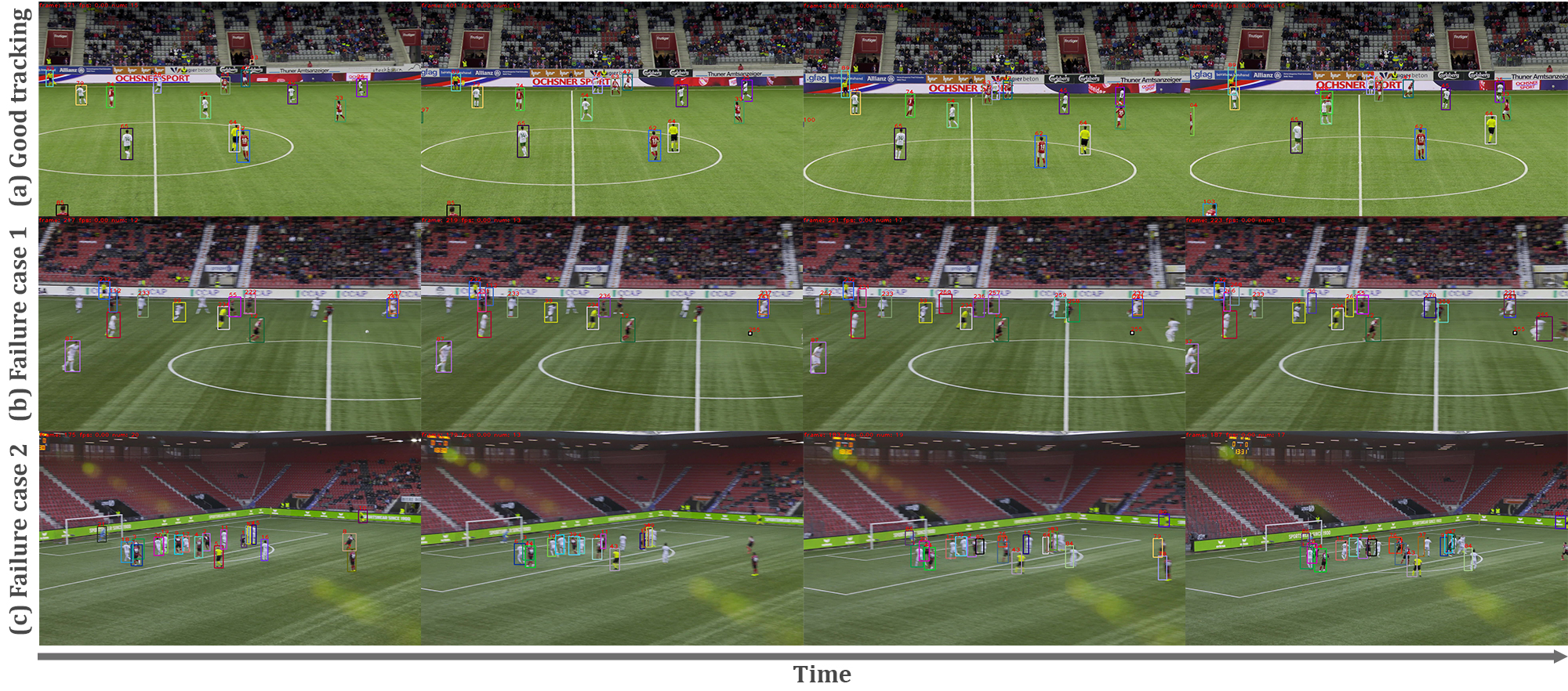}
    \caption{
    \textbf{Qualitative tracking results.} Tracking sequences produced by ByteTrack with ground-truth detections. Sequence (a) represents a good tracking of the players, even after some players or the referee are partially occluded.
    Sequence (b) shows an example of challenging association due to fast motion of the ball and players between consecutive frames.
    Sequence (c) displays a challenging free-kick scenario where many players are clustered together resulting in extreme occlusions and poor tracking results.} 
    \label{fig:qualitative_tracking}
\end{figure*}

\begin{figure}[t]
    \centering
    \includegraphics[width=\linewidth]{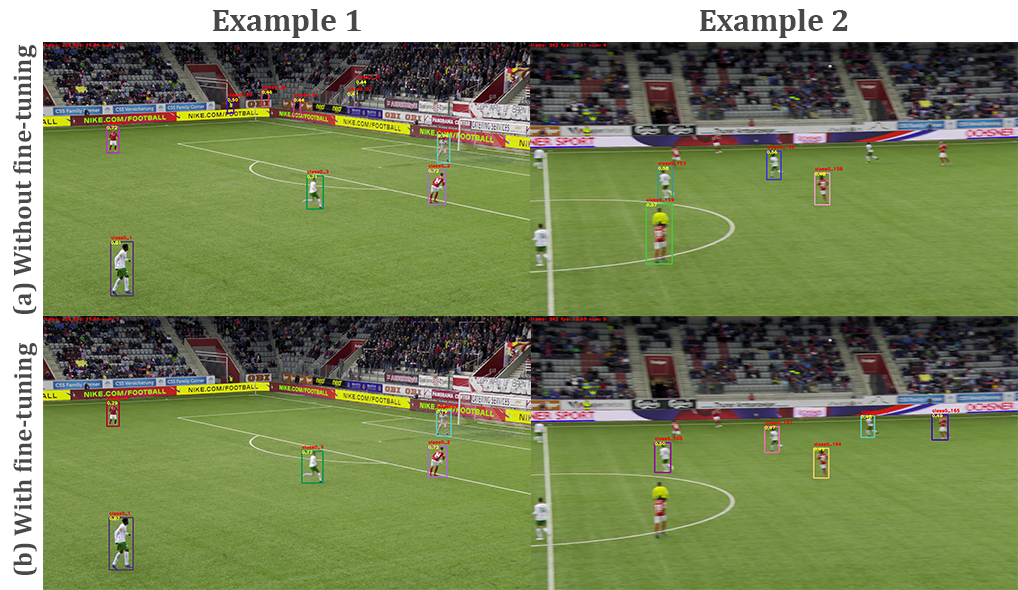}
    \caption{
    \textbf{Qualitative fine-tuned results.} 
    Comparison of FairMOT predictions before and after fine-tuning on our dataset in the setup ``w/o GT''. 
    Example 1 shows that the fine-tuned model does not consider the audience anymore in its predictions.
    Example 2 shows that we can also better detect player bounding boxes, despite the fact that fast motion blur remains challenging.
    }
    \label{fig:qualitative_finetune_results}
\end{figure}

Second, \emph{when ground-truth detections are not provided}, we can observe increasingly better performance from DeepSORT to FairMOT to ByteTrack, which is aligned with other MOT benchmarks.
Furthermore, we note that FairMOT has better detection capabilities (DetA) than other baselines while ByteTrack has better association capabilities (AssA). Consequently, FairMOT has a better MOTA as this metric favors detection over association performance.

Third, we can see that fine-tuning FairMOT on our data improves both the detection and association performance, leading to state-of-the-art scores for both HOTA and MOTA.
This improvement mainly comes from two sources: (i) The original detectors for the three baselines are trained to detect all humans in the video, hence including the audience, whereas our annotations only select players or staff on the field. Therefore, this results in lots of false positive detections outside the field that significantly lower the detection score. (ii) The original detectors are only trained to detect humans and hence never detect the ball which is considered in our annotations. Fine-tuning our model to only consider the humans and the ball on the field leads to less false positives and false negatives in the detections and therefore improves both the HOTA and MOTA scores.
 
\mysection{Per-class analysis.} Furthermore, we provide a thorough performance study per action class of the ByteTrack baseline in Table~\ref{table:hota_categories} for our two setups.
\emph{When ground-truth detections are provided}, detection scores are significantly lower for actions involving fast motion, mostly due to the filtering from the association process as explained above. For instance, substitutions are mostly static events and therefore have a high DetA, while clearances and goals involve fast moving players leading to lower DetA scores.
This is also reflected in the overall HOTA score, for which the easiest classes are substitutions and yellow cards. Interestingly, the hardest cases are corners, which may be explained by lots of ID switches, as reflected by AssA, with players crossing each other multiple times during those scenarios.
 
Last, \emph{when ground-truth detections are not provided}, substitutions and yellow cards remains the easiest scenarios to resolve, meaning that the generic detector already does a pretty good job in tracking the players in those scenarios. Corners and fouls remain challenging in this setup, but the two hardest scenarios are now goals and penalties, which often involve challenging dense player clusters such as celebrations or walls of players. 
In those cases, object detectors perform poorly due to players overlaps and occlusions.
Indeed, we can see that when detections are given, these classes are actually well tracked, indicating that the difficulty is rather in detecting the objects in these scenarios than associating the correct bounding boxes together.

\mysection{Qualitative results.} We show three different sequences for a qualitative analysis of the ByteTrack in Figure~\ref{fig:qualitative_tracking}. 
In sequence (a), we can see that the players are well tracked even after some players or the referee partially occlude each other. 
Sequence (b) shows an example of challenging association due to fast motion of the ball and players between consecutive frames.
Sequence (c) shows a free-kick scenario, where many players are clustered, resulting in extreme occlusions and ByteTrack performing very badly. Our hope is that future methods have better re-identification capabilities to resolve hard cases like these.

Finally we analyse the effect of fine-tuning FairMOT on two frames in Figure~\ref{fig:qualitative_finetune_results}.
In the first example, we can see the inability of the pre-trained detector to distinguish between people in the audience and players on the field. In contrast, the fine-tuned detector has learned the semantic roles of each actor, discarding the audience. The second example shows a harder case with fast moving objects, where the fine-tuned detector learns to detect some of the fast moving players but not all of them. This indicates that further work is needed to improve detection performance as well.

%

\vspace{-2mm}
\section{Conclusion}
\label{sec:Conclusion}
\vspace{-2mm}

We release the novel SoccerNet-Tracking dataset, which is the largest dataset for multi-object tracking in soccer, featuring challenges such as high-speed movements and highly occluded objects. This topic is important for many research and industrial application, which can have a direct impact on soccer. As a first approach, we study three state-of-the-art methods for multi-object tracking and discuss the challenging situations in our dataset that are not correctly tackled by those methods in their current state. With SoccerNet-Tracking, we have set a first benchmark and aim at pushing the computer vision community towards better tracking methods, including long-term re-identification in challenging environments, by organizing tracking challenges.

\mysection{Acknowledgement.}
This work was supported by the Service Public de Wallonie (SPW) Recherche, under Grant No. 2010235 -- ARIAC by \href{http://DigitalWallonia4.ai}{DigitalWallonia4.ai}, and KAUST Office of Sponsored Research (CRG2017-3405).
\clearpage


\end{document}